\documentclass[pmlr]{jmlr}% new name PMLR (Proceedings of Machine Learning Research)

\usepackage{longtable}% for long tables

\usepackage{booktabs}
\usepackage{siunitx}
 
\usepackage[utf8]{inputenc}
\usepackage{graphicx}
\usepackage{color}
\usepackage{hyperref}
\usepackage{wrapfig}

\newcommand{\namenospace}{MineRL \textsc{BASALT}}
\newcommand{\name}{MineRL \textsc{BASALT}\ }
\newcommand{\diamondnamenospace}{MineRL Diamond}
\newcommand{\diamondname}{MineRL Diamond\ }

\usepackage{cleveref}

\jmlrvolume{}
\jmlryear{}
\jmlrworkshop{PMLR NeurIPS 2021 Demo \& Competition Track Volume}

\title[\name Retrospective]{Retrospective on the 2021 \name Competition \titlebreak on Learning from Human Feedback}
 
 \author{
 \Name{Rohin Shah} \Email{rohinmshah@berkeley.edu} \\
 \addr University of California, Berkeley
 \AND
 \Name{Steven H. Wang} \Email{sh.wang@berkeley.edu} \\
 \addr University of California, Berkeley
 \AND
 \Name{Cody Wild} \Email{cody.marie.wild@gmail.com} \\
 \addr University of California, Berkeley
 \AND
 \Name{Stephanie Milani} \Email{smilani@cs.cmu.edu} \\
 \addr Carnegie Mellon University
 \AND
  \Name{Anssi Kanervisto} \Email{anssk@uef.fi} \\
 \addr University of Eastern Finland
 \AND
 \Name{Vinicius G. Goecks} \Email{vinicius.goecks@gmail.com} \\
 \addr DEVCOM Army Research Laboratory
 \AND
 \Name{Nicholas Waytowich} \Email{nicholas.r.waytowich.civ@army.mil} \\
 \addr DEVCOM Army Research Laboratory
 \AND
 \Name{David Watkins-Valls} \Email{davidwatkins@cs.columbia.edu} \\
 \addr Columbia University
 \AND
 \Name{Bharat Prakash} \Email{bhp1@umbc.edu} \\
 \addr University of Maryland, Baltimore County
 \AND
 \Name{Edmund Mills} \Email{edmund@edmundmills.com} \\
 \addr Independent
 \AND
 \Name{Divyansh Garg} \Email{divgarg@stanford.edu} \\
 \addr Stanford University, Stanford
 \AND
 \Name{Alexander Fries} \Email{Alexander.Fries27@gmail.com} \\
 \addr Independent
 \AND
 \Name{Alexandra Souly} \Email{alexandrasouly@gmail.com} \\
 \addr Independent
 \AND
 \Name{Chan Jun Shern} \Email{chanjunshern@gmail.com} \\
 \addr Independent
 \AND
 \Name{Daniel {del Castillo}} \Email{ddcastilloiglesias@gmail.com} \\
 \addr Independent
 \AND
 \Name{Tom Lieberum} \Email{tlieberum@outlook.de} \\
 \addr University of Amsterdam
 }

\editor{}
\begin{document}

\maketitle
\newpage

\begin{abstract}
We held the first-ever MineRL Benchmark for Agents that Solve Almost-Lifelike Tasks (\namenospace) Competition at the Thirty-fifth Conference on Neural Information Processing Systems (NeurIPS 2021). The goal of the competition was to promote research towards agents that use learning from human feedback (LfHF) techniques to solve open-world tasks. Rather than mandating the use of LfHF techniques, we described four tasks in natural language to be accomplished in the video game Minecraft, and allowed participants to use any approach they wanted to build agents that could accomplish the tasks.

Teams developed a diverse range of LfHF algorithms across a variety of possible human feedback types. The three winning teams implemented significantly different approaches while achieving similar performance. Interestingly, their approaches performed well on \emph{different} tasks, validating our choice of tasks to include in the competition.

While the outcomes validated the design of our competition, we did not get as many participants and submissions as our sister competition, \diamondnamenospace. We speculate about the causes of this problem and suggest improvements for future iterations of the competition.
\end{abstract}

\section{Introduction} \label{sec:introduction}

Specifying reward functions for AI systems by hand can often lead to \emph{specification gaming}, in which the AI agent finds a way to achieve high reward without behaving how the designer intended~\citep{krakovna2018specification, lehman2018surprising, kerr1975folly}. In an infamous example, an agent trained to play the boat racing game CoastRunners discovers it can loop forever collecting power-ups that give it bonus points: significantly increasing the score, but never finishing the race~\citep{clark2016faulty}.

Specification gaming occurs because the reward function is interpreted as defining the optimal behavior of the agent in every situation that can arise in the environment, even though in practice it should be treated as a potentially flawed source of evidence about what the task designer wants~\citep{hadfield2017inverse}. We thus aim to incorporate additional channels for communicating information about what designers want to our agents. We call the class of techniques for this problem ``learning from human feedback'' (LfHF). \citet{jeon2020reward} provide a survey of possible feedback modalities, including demonstrations~\citep{ng2000algorithms, ziebart2010modeling}, comparisons~\citep{christiano2017deep, wirth2017survey}, the state of the world~\citep{shah2019preferences}, and more. Note that LfHF includes imitation learning as a special case where the human feedback is given through demonstrations.

To promote research towards agents that learn from human feedback rather than a predefined reward function, we held the first-ever MineRL Benchmark for Agents that Solve Almost-Lifelike Tasks (\namenospace) Competition at the 35th Conference on Neural Information Processing Systems (NeurIPS 2021). To our knowledge, \name is the first competition to test LfHF techniques in a sequential decision-making setting. Teams developed agents for four open-world tasks in Minecraft: finding a cave, building a scenic waterfall, corralling animals without disrupting a village, and safely building a village house. We provided a dataset of human demonstrations of each task, using the same infrastructure as in MineRL Diamond~\citep{guss2019neurips}. Our challenge attracted approximately 300 registered participants, of which 10 teams provided submissions to all of the required tasks.

\begin{figure}[ht]
    \centering
    \includegraphics[trim=0 160 0 0,clip,width=\textwidth]{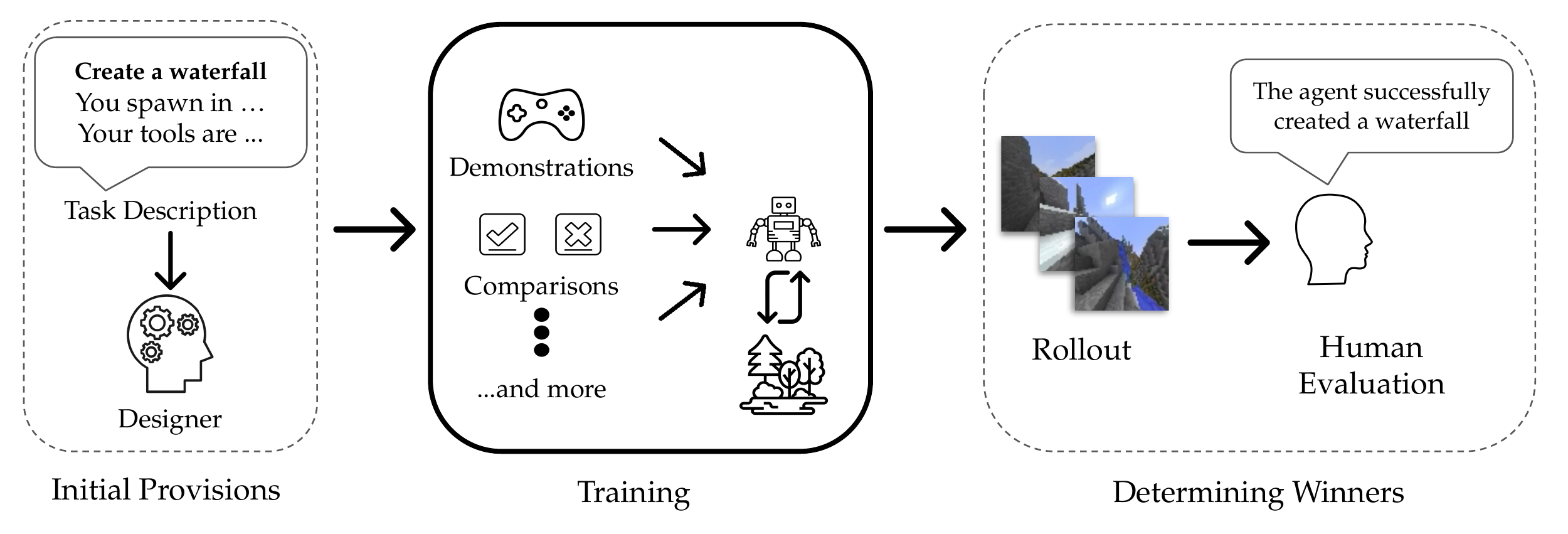}
    \caption{An illustration of the \name competition procedure.}
    \label{fig:overview}
\end{figure}

\section{Competition Overview} \label{sec:overview}

We provide a brief overview of the \name competition; for a full description of the motivation, design, and implementation, we refer the reader to the competition proposal~\citep{shah2021minerl} and website~\citep{shah2021minerl_aicrowd}.

Figure~\ref{fig:overview} shows the overall competition structure. We provided tasks consisting of a simple English language description alongside a Gym environment, \emph{without} any associated reward function (Section~\ref{sec:tasks}). Participants trained agents for these tasks using any methods of their choice. The submitted agents were evaluated based on how well they completed the tasks, as judged by humans given the same task descriptions (Section~\ref{sec:evaluation}).

\subsection{Tasks} \label{sec:tasks}

Our goal was to promote work on LfHF algorithms that can eventually be applied to hard-to-specify problems of interest in the real world. Since we wanted algorithms to transfer to real-world settings, we wanted to incentivize general solutions rather than algorithms that are specific to a given task. This drove our choice of Minecraft as our environment. Minecraft is extremely open-ended: humans pursue a wide variety of goals within the game, and the game mechanics provide a hierarchical and combinatorial space of possible tasks that agents can be trained to perform. By designing tasks involving high-level concepts that cannot be easily identified programmatically, we were able to recreate the situation we often face in real-world settings: reward functions cannot be used simply because they are too hard to write down.

We designed a suite of four tasks that participants were required to design performant agents for: FindCave, MakeWaterfall, CreateVillageAnimalPen, BuildVillageHouse. Participants submitted a separate trained agent along with the corresponding training code for each task. Crucially, the training code could expose an English-language interface (either via the command line, or some more complex GUI if desired) where humans with at most 30 minutes of training could provide feedback. We did not require participants to use identical training code across all four tasks, but we hoped that having performance on all tasks count towards the final competition score would encourage the use of general-purpose techniques rather than task-specific hard-coding.

\subsection{Evaluation} \label{sec:evaluation}

For each task, we executed submitted agents on test environments to produce multiple videos of agents attempting to complete that task. We then compare agents to each other by putting them in ``matches''. In each match, we choose two agents, one task, and one test seed, and show the two corresponding videos to a human contractor. The contractor then determines which agent better completed the task. Given a dataset of such comparisons, we used the TrueSkill system~\citep{herbrich2006trueskill} to compute scores for each agent. Since we expected agent skills to vary significantly across tasks, we computed separate TrueSkill scores for each task. We collected comparisons until the scores produced a reasonably stable ranking of agents, ultimately requiring thousands of comparisons. The standard deviation estimated by TrueSkill for an individual submission was typically around $0.7$.

To determine winners, we need to aggregate TrueSkill scores across tasks. One intuitive way to do this is to normalize the TrueSkill scores \emph{across submissions to the task}, by transforming a score of $x$ to $\frac{x - \mu}{\sigma}$, where $\mu$ is the mean and $\sigma$ is the standard deviation of the scores. (Note that here $\mu$ and $\sigma$ are \emph{not} the means and standard deviations estimated by TrueSkill, but are rather the mean and standard deviation calculated across participant scores.) We can then take the average of the normalized scores to determine an overall score. The normalization ensures that no one task unduly drives the final scores, while they remain sensitive to the magnitude of the differences in scores between different agents on a task. Unfortunately, this scheme has a major downside: if all submissions behave similarly on a task, then $\sigma$ will be very small, and tiny differences between agents can be amplified into large differences in overall score. To mitigate this issue, we required the denominator to be at least 1, such that a score of $x$ was mapped to $\frac{x - \mu}{\max(\sigma, 1)}$.

For the top submissions in contention for winning a prize, we looked over the submitted training code to ensure that rules were not broken. We then retrained the agents with a budget of up to four days of compute and ten hours of online human feedback. Participants were allowed to choose how to split their compute and human feedback budget across the four tasks. We checked that the retrained agents behaved similarly to the originally submitted agents. If there were significant discrepancies, we would have disqualified the submission, but luckily this did not prove to be necessary.

\subsection{Resources}

\paragraph{AIcrowd website.} AIcrowd provided a unified interface for participants to sign up for the competition, submit their trained agents, ask the organizers and other participants questions, and monitor their progress on a public leaderboard.

\paragraph{Baselines.} We provided a behavioral cloning (BC) baseline to participants to help them get started on their solution.

\paragraph{Mentorship.} Both \name and \diamondname used an existing MineRL Discord server to create an active community to work on the competition problems. This was very helpful in increasing engagement and helping solve any issues that arose.

\paragraph{Computing and evaluation resources.} We offered \$500 compute grants to participants that self-identified as lacking access to the necessary resources to take part in the competition. We provided these grants to 26 out of the 27 applications we received.

\paragraph{Tutorial and Documentation.} We provided a website that contains instructions, documentation, and updates to the competition. We also provided references to prior LfHF approaches that participants could read for ideas on how to tackle the competition.

\subsection{Related Competitions}

To our knowledge, \name was the first competition to test LfHF techniques in a sequential decision-making setting. Our sister competition, MineRL Diamond~\citep{guss2019neurips}, also leverages human demonstrations, but does so in order to improve sample efficiency on tasks with a fixed reward function, whereas we focus on using human demonstrations (and other human feedback) to \emph{replace} the reward function. There are also several other competitions set in Minecraft~\citep{perez2019multi, gray2019craftassist, johnson2016malmo, salge2018generative}, though their focuses are quite different from ours.

\section{Winning Submissions} \label{sec:winners}

Having described the competition, we now turn to the winners and their approaches.

\begin{table}[bhtp]
\floatconts
  {tab:example-siunitx}
  {\caption{{\bfseries Leaderboard.} We show the scores that teams achieved on each task, as well as the average (used to determine winners). The scores are TrueSkill scores that have been normalized according to the method described in Section~\ref{sec:evaluation}. `Baseline'' is the behavioral cloning (BC) method that we provided at the start of the competition.}}
  {\begin{tabular}{lSSSSS} \label{table:winners}
  \\ \toprule
  \bfseries Team & {\bfseries FindCave} & {\bfseries Waterfall} & {\bfseries AnimalPen} & {\bfseries House} & {\bfseries Average}\\
  \midrule
  KAIROS & -0.23 & {\bfseries 2.81} & 0.15 & -0.06 & {\bfseries 0.67} \\
  obsidian & {\bfseries 1.07} & 0.21 & 1.00 & 0.15 & {\bfseries 0.61} \\
  NotYourRL & 0.44 & 0.13 & {\bfseries 1.59} & 0.03 & {\bfseries 0.55} \\
  mina & 0.80 & -0.71 & -0.08 & {\bfseries 0.18} & {\bfseries 0.05} \\
  yamato.kataoka & -0.17 & 0.00 & -0.24 & 0.00 & {\bfseries -0.10} \\
  Reforcos\_de\_Minecraft & -0.60 & 0.18 & -0.25 & -0.03 & {\bfseries -0.17} \\
  UEF & -0.19 & -0.49 & -0.03 & -0.05 & {\bfseries -0.19} \\
  Baseline & -0.26 & -0.43 & -0.04 & -0.12 & {\bfseries -0.21} \\
  chrischongtt & -0.60 & -0.32 & -0.20 & -0.06 & {\bfseries -0.30} \\
  Granite & 0.10 & -0.61 & -1.12 & 0.00 & {\bfseries -0.41} \\
  PA-P & -0.36 & -0.76 & -0.79 & -0.05 & {\bfseries -0.49} \\
  \bottomrule
  \end{tabular}}
\end{table}

\subsection{Prize Winners}

We show the leaderboard in Table~\ref{table:winners}. The first, second, and third place prizes went to Teams KAIROS, obsidian, and NotYourRL, respectively. Besides the main prizes for the top three teams on the leaderboard, we had two other prizes for submissions and two smaller prizes to encourage activity in community:

\paragraph{Most human-like prize.} This prize went to the submission with the most human-like behavior. This was judged in exactly the same way as the overall leaderboard, except that instead of having human raters compare agents based on how well they completed the task, we had them compare agents based on how human-like their behavior was. Team KAIROS, the overall first-place winner, also won this prize.

\paragraph{Creativity of research prize.} This prize went to the submission whose approach was the most creative, as judged by the organizing committee. Team NotYourRL, the overall third-place winner, also won this prize.

\paragraph{Community support prize.} We awarded three Discord users with a prize for answering participant questions and discussing the competition, rules, and potential solutions.

\paragraph{Evaluation lottery prize.} We gave out five lottery awards for those who submitted human evaluations for the online leaderboard to encourage people to do the online evaluations.

\begin{figure}[!ht]
  \centering
  \includegraphics[width=\linewidth]{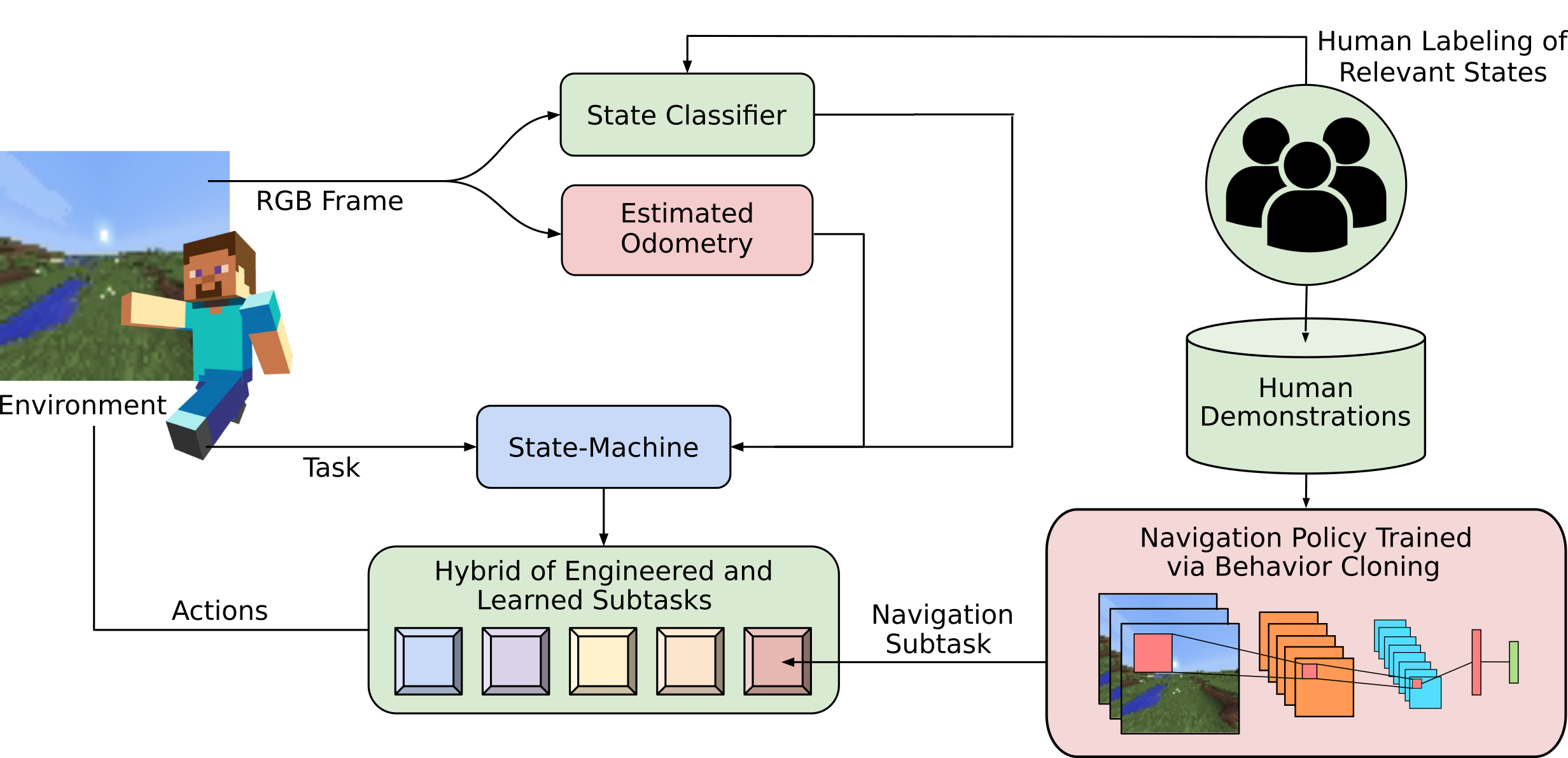}
  \caption{Diagram illustrating the KAIROS team approach~\citep{goecks2021combining}.}
  \label{fig:kairos_diagram}
\end{figure}

\subsection{First Place: KAIROS} \label{sec:kairos}

The KAIROS team opted for a hybrid intelligence approach where subtasks, either engineered by human designers or learned from human feedback data, are hierarchically orchestrated by a state-machine~\citep{goecks2021combining}.

As seen in their main approach diagram, shown in Figure~\ref{fig:kairos_diagram}, the transitions between the states of the state-machine are controlled by two main modules: a state classifier and an estimated odometry module.
The state classifier consists of a convolutional neural network that takes the agent's camera view of the environment as input and classifies it in one of twelve possible classes such as if the image contain caves, mountains, animals, water, and others. This classifier was trained using approximately $82,000$ human-labeled images.
The estimated odometry module assumes the agent follows point-mass kinematics and estimates its motion based on the actions taken by the agent and mechanics of the Minecraft simulator, such as that the time interval between each frame is $0.05$ seconds and that the terminal walking speed on Minecraft\footnote{Minecraft Wiki - Walking: \url{https://minecraft.fandom.com/wiki/Walking}.} is approximately $4.317$ m/s.

The multiple subtasks controlled by the state-machine are either learned from human data or engineered by human designers.
The main subtask is a navigation policy that drives the agent to key locations in the environment such as the ideal spot to place a waterfall, or start the construction of the pen or house.
This policy is a convolutional neural network trained via imitation learning using the human demonstration dataset provided by the competition after it was preprocessed to mask out all non-movement-related actions.
The remaining subtasks were designed by a human expert to use the output of the learned state classifier and estimated odometry model to perform behaviors that were not able to be learned directly from data. The main example is the ``end episode" subtask for the \textit{FindCave} task, where the agent is scripted to equip and throw the snowball whenever the state classifier is confident that the agent is inside a cave.

\begin{wrapfigure}{R}{0.5\linewidth}
    \vspace{-12pt}
  \begin{center}
  \includegraphics[width=0.9\linewidth, trim=0 24pt 0 24pt]{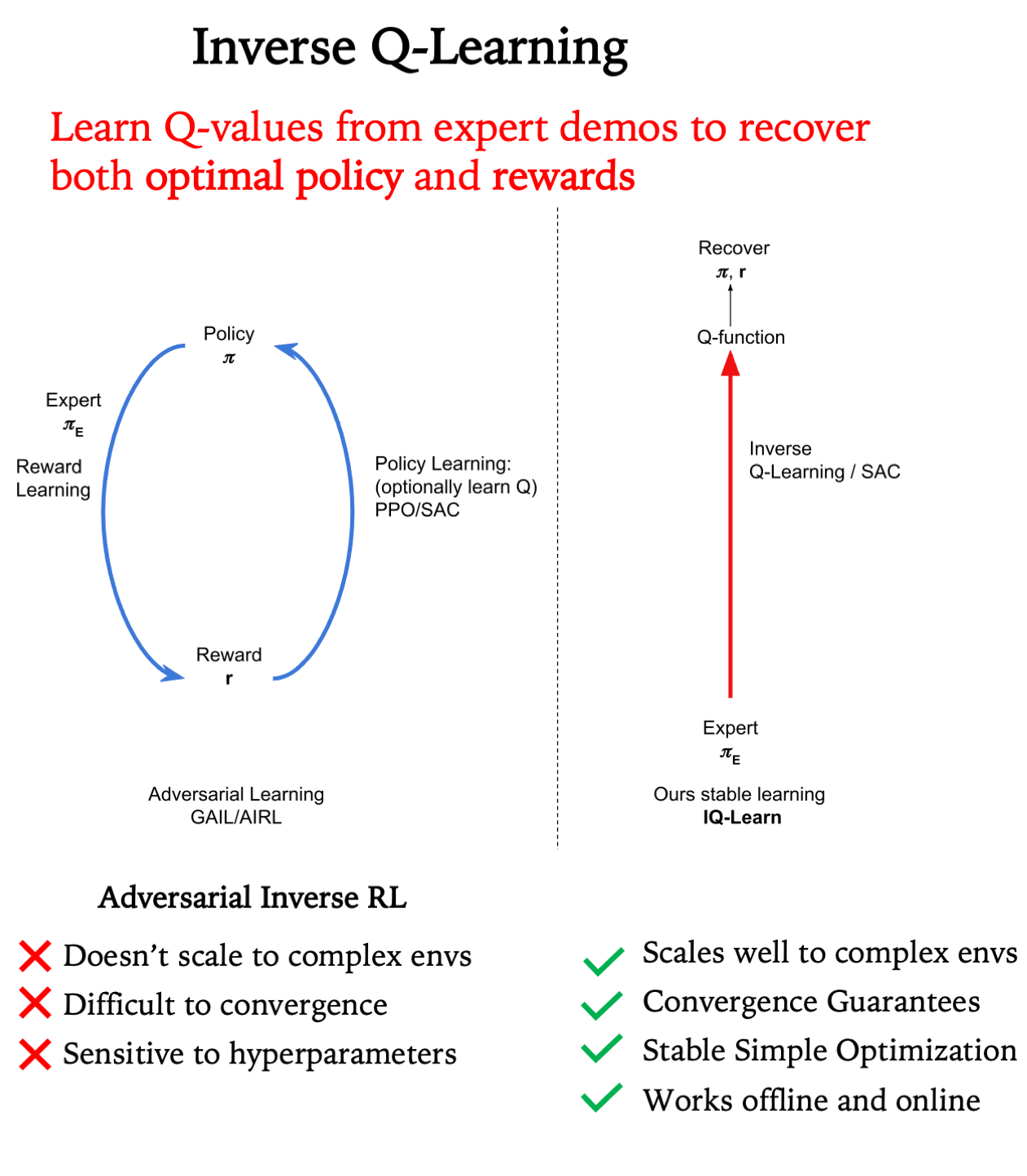}
  \end{center}
     \vspace{-24pt}
    \caption{IQ-Learn method.}
    \label{fig:obsidian}
     %\vspace{-12pt}
\end{wrapfigure}

\subsection{Second Place: Obsidian} \label{sec:obsidian_approach}

The Obsidian team's approach used a novel state-of-art algorithm for imitation learning: \textbf{Inverse Q-Learning (IQ-Learn)\footnote{IQ-Learn link: \url{https://div99.github.io/IQ-Learn}.}}~\citep{garg2021iqlearn} to train an agent on each of the tasks. Their approach did not use any human-provided interactions, auxiliary data, or labels and relied on pure imitation of the given offline demonstrations. A crucial part of their approach was using a Recurrent Neural Network (RNN) as part of the agent's policy.

IQ-Learn is the first non-adversarial method for Inverse Reinforcement Learning (IRL), developed using new theoretical insights. It works with sparse expert data and scales to complex environments, improving on prior works like GAIL~\citep{ho2016generative} and SQIL~\citep{reddy2019sqil} by orders of magnitude. IQ-Learn learns a Q-function using human demonstrations, which implicitly represents both a reward function and a policy. It optimizes a policy using a single training loop, in contrast to existing inverse reinforcement learning and adversarial imitation learning, which use nested training loops and alternative updates and are more complex to implement. At the same time, in contrast to simpler methods like behavioral cloning, the policy learned using IQ-Learn can represent the environment's dynamics and enables online imitation learning. In this competition, team Obsidian used the online version of IQ-Learn, which trains agents using experiences sampled from both human demonstrations and a replay buffer generated by environment interaction.

The agent's policy contained an LSTM~\citep{hochreiter1997long}, which allowed it to learn more complex behaviors than similarly trained non-recurrent policies. Examples of such behavior include swimming, which requires performing the same action repeatedly over many timesteps. Additionally, in the MakeWaterfall task, the LSTM enabled the agent to learn to stay nearby and end the task after placing the waterfall rather than moving away. This behavior was a significant differentiator of success and failure on the task.

The same approach was used for each of the competition's tasks, demonstrating the generality of this approach. Between each task, only the following varied:
\begin{itemize}
\item The discretization of the action space to allow the use of environment-specific items.
\item The vectorization of the inventory and equipped item components of the observations.
\item A few hyperparameters, most significantly the model size and experience sequence length used to train the LSTM. The MakeWaterfall task particularly benefited from a longer sequence length.
\item Training duration, based on perceived task difficulty.
\end{itemize}

This approach used relatively little domain knowledge: the action space was discretized roughly following one of the provided baselines; all no-op actions were removed from the expert demonstration dataset; the demonstrations were augmented using random horizontal shifts of the image component of the observation; and the agent was only allowed to terminate an episode if its likelihood of doing so exceeded a confidence threshold, to reduce its tendency to end the episode early.

\subsection{Third Place: NotYourRL} \label{sec:notyourrl}

Team NotYourRL's approach was mostly based on the approach in~\citet{ibarz2018reward}. At a high level, this constitutes two main parts: 
learning Q-values using Deep Q-learning from Demonstrations~\citep{hester2018deep} and learning the reward model from human preferences~\citep{christiano2017deep}.

DQfD uses double Q-learning with prioritized experience replay \citep{schaul2015prioritized} that allows for prioritizing human demonstrations over agent trajectories, while learning from both. As deep Q-learning needs a reward signal for its training, the reward model is trained using sample clips from demonstrations and agent trajectories. This reward model, a convolutional neural network (CNN), then estimates the corresponding reward function from elicited trajectory preferences and expert demonstrations.  

Due to time constraints, the team used an auto-labeling approach to generate preferences over clips instead of using human judgment. Clips are randomly sampled from the trajectory buffer and are labeled as follows: Clips from human demonstrations are always preferred to agent-generated clips. Agent-generated clips are always preferred to random-policy clips. Finally, clips from later in the episode are preferred to ones earlier in the episode on the assumption that the reward is mostly accrued towards the end of an episode.

The reward model is set up to predict a reward for each clip, so that the probability of a clip being preferred is higher when the reward is higher. It is trained using cross-entropy loss to minimize the disparity between the predicted and the actual preference between clip pairs. As training progresses, the number of annotated pairs that are collected decreases. At the beginning of training, no annotator is needed as clips of the initial trajectories are paired with expert demonstration clips (where the latter is always preferred).

\begin{wrapfigure}{R}{0.35\linewidth}
    \vspace{-12pt}
  \begin{center}
  \includegraphics[width=0.85\linewidth]{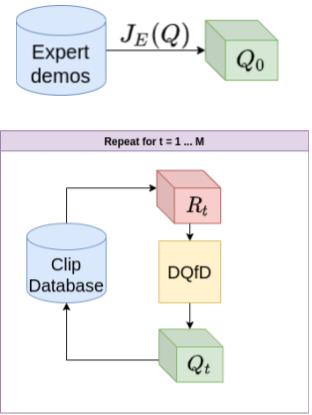}
  \end{center}
    \caption{Diagram describing Team NotYourRL's approach.}
    \label{fig:notyourrl_diag}
    \vspace{-60pt}
\end{wrapfigure}

A simplified structure of the algorithm can be seen in \cref{fig:notyourrl_diag}. First, a Q-net is trained via imitation learning using expert classification loss. Then, for $M$ iterations of the training loop,
\begin{itemize}
    \item Preferences are collected over sample clips.
    \item The reward model is trained using these preferences.
    \item The Q-network is trained using the reward model.
    \item New trajectories are collected using the Q-network.
\end{itemize}

 A surprising finding was that while the reward model was able to predict the preferred clip with an accuracy over 80\% in all four tasks, the reward signal did not improve the observed agent behavior visibly. Possibly, while the negative and positive examples were distinct enough, they did not confer enough information about the underlying task. Therefore, the model trained on the reward model was not included in the final submission, only the model using the Q-network trained via imitation learning. Using real human feedback would most likely improve the impact of the reward model on the agent's behavior, which suggests clear paths to improvement for future research.
 
 \paragraph{Action shaping.} In order to reduce the size of the action space, team NotYourRL discretized camera actions (and applied thresholds on human camera actions), discarded nonsensical combinations (e.g., forward + backward), discarded sneak and sprint, and allowed only one choice per action ``group'' (e.g., items, camera actions, etc).

\section{Reflections}
The results of the competition have shown both that many aspects of the competition worked well, and that there are many areas on which to improve.

\subsection{Task Design}

We designed the four tasks to cover a wide variety of possible scenarios, and to provide different levels of difficulty. FindCave was meant to be the easiest task, while BuildVillageHouse was meant to be the hardest. In line with this, not much progress was made on BuildVillageHouse, with all submissions receiving roughly similar scores. However, even FindCave was hard to solve: while the top submissions clearly outperformed the baseline, they were far from completing the task successfully as often as humans do.

We were also pleased to see that each of the top three teams did well on a different task (Table~\ref{table:winners}). This suggests that our tasks are capturing a variety of different requirements that are traded off between approaches. In addition, these teams used significantly different techniques, which serves the research goals of this competition.

\subsection{Evaluation Methodology}

One notable feature of this competition was the use of human comparisons to \emph{evaluate} trained agents. This allowed us to stick closely to the real-world setting, in which there is no programmatic specification of what an AI system has to do. One benefit we noticed was that this let us have a very relaxed set of rules, since any method of solving the task would count as progress, and indeed we had barely any questions about rule clarifications from participants, unlike other competitions. In addition, using human comparisons can allow for more granular feedback, such as descriptions of \emph{how} agents tend to fail.

However, this methodology does come with downsides. When agents behave relatively similarly, there can be a lot of noise in the results. In this particular competition, while there was a large difference between the top three teams and all other teams, there was only a small difference between the top three teams. It is plausible that running the evaluation again would lead to a different ranking of the top three teams. In addition, since the scores and rankings were computed from comparisons, they do not have meaning independent of the group of agents that were being compared. This means that one cannot easily compare scores across different iterations of the competition.

\subsection{Organizational Details}

Despite some technical hiccups from building new evaluation systems, the competition was executed relatively smoothly. Our community engagement with the MineRL Discord server was particularly well-received, and we received positive comments regarding how active the community and organizers were in discussing and answering questions. We also received messages thanking us for organizing the competition as it encouraged some of the participants to learn about programming and machine learning.

The main organizational issue was that our timeline for evaluation was too short. Despite having fewer submissions than we expected, and none of the top teams using online human feedback (making retraining much simpler), we were only able to complete evaluation and announce winners about a week before the competition workshop. While future iterations of the contest would have better-tested systems, we would still recommend leaving three months between the submission deadline and the competition workshop.

\subsection{Participation}

Despite active advertising across a variety of forums, including the BAIR blog\footnote{\url{https://bair.berkeley.edu/blog/2021/07/08/basalt/}}, Twitter, and Hacker News, as well as signal boosts from VentureBeat\footnote{\url{https://venturebeat.com/2021/07/09/basalt-minecraft-competition}} and Yannic Kilcher\footnote{\url{https://www.youtube.com/watch?v=-cT-2xvaeks&t=715s}}, this competition had fewer participants than related competitions. MineRL Diamond had around 450 registered participants and 1190 submissions, while BASALT had around 300 registered participants and 271 submissions. Only about five teams used unique methods instead of modifying or extending the baseline behavioral cloning code. While it is hard to know the exact causes, we suspect at least the following contributors:
\begin{enumerate}
    \item The submissions system was difficult to use, particularly because we had four subtasks. We also received comments that the landing page for the competition was confusing.
    \item The unique challenge of this competition, learning from human feedback, is hard to approach. In particular, it is challenging to evaluate agents due to the lack of a reward function, making it hard to judge whether a particular change was good.
    \item The online public leaderboard did not have enough evaluations for a reliable ranking, meaning that participants did not receive good intermediate feedback. The final leaderboard (\Cref{table:winners}) had a completely different ranking to the online leaderboard.
    \item Even the ``easy'' task (FindCave) was challenging, making it hard to iterate quickly.
\end{enumerate}

\subsection{Recommendations for Future Iterations}

We would recommend that future iterations of the competition make the following changes:
\begin{enumerate}
    \item If using the same competition structure, allow at least 3 months for evaluation.
    \item Make the public leaderboard accurately track participant rankings, or remove it.
    \item Lower the barrier to entry to participate. Create additional baselines and provide better tutorials and easier installation. This would also increase the diversity of solutions from teams that decide to modify a baseline.
\end{enumerate}

\section{Conclusion}
We ran the MineRL BASALT Competition at NeurIPS 2021 to promote the development of agents that learn from human feedback. We believe that the competition succeeded in its main goals, with participants creating multiple different solutions that significantly improved upon the baseline. However, there are many avenues for improvement in order to run an even more successful competition in the future.

\acks{
This competition would not have been possible to run without the help of many people and organizations. Open Philanthropy, Microsoft, AI Journal, OpenAI, and Google provided financial support. Sharada Mohanty, Shivam Khandelwal, Vrushyank Vyas, and other AIcrowd employees created submission and evaluation systems for participants to use. Prashanth, Sravan, Soumya, Ramesh, Sahil, Ram, Jaswantsinh, and Sumit provided the comparisons used to evaluate participant submissions and determine winners.
}

\bibliography{references}

\end{document}